%% file: main.tex
\documentclass[11pt,a4paper]{article}
\usepackage{times}
\usepackage{float}
\usepackage{url}
\usepackage{comment}
\usepackage{latexsym}
\usepackage{amsmath}
\usepackage{graphicx}
\usepackage{xcolor}
\usepackage{comment}
\usepackage{url}
\usepackage{multicol}
\usepackage{booktabs}
\usepackage{colortbl}
\usepackage{subcaption}
\usepackage{caption}
\usepackage[colorinlistoftodos]{todonotes}
\usepackage{natbib}
\usepackage[margin=25mm]{geometry}
\usepackage{dirtytalk}

\newcommand{\cn}{\textsc{ConceptNet}}

\newcommand{\rel}[1]{\textsc{#1}}

\newcommand{\tr}[3]{$\langle${\it #1},\rel{#2},{\it #3}$\rangle$}

\title{Assessing the Difficulty of Classifying ConceptNet Relations \\in a Multi-Label Classification Setting}

\author{Maria Becker, Michael Staniek, Vivi Nastase, and  Anette Frank \\
Heidelberg University, Department of Computational Linguistics \\
{\tt \{mbecker/staniek/nastase/frank\}@cl.uni-heidelberg.de} \\ \\}

\date{}

\begin{document}
\maketitle

\thispagestyle{empty}
\pagestyle{empty}

\begin{abstract}
Commonsense knowledge relations are crucial for advanced NLU tasks. We examine the learnability of such relations as represented in \cn, taking into account their \textit{specific properties}, which can make relation classification difficult: a given concept pair can be linked by multiple relation types, and relations can have multi-word arguments of diverse semantic types. We explore a neural \textit{open world multi-label classification approach} that focuses on the evaluation of 
 classification accuracy for individual relations. Based on an in-depth study of the specific properties of the \cn~resource, we investigate the impact of different relation representations and model variations. Our analysis reveals that the complexity of argument types and relation ambiguity are the most important challenges to address. We design a customized evaluation method to address the incompleteness of the resource that can be expanded in future work.

\end{abstract}

\input{1Intro.tex}

\input{2RelWork.tex}

\input{4RelClass.tex}

\input{5Experiments.tex}

\input{5results.tex}

\input{6Analysis.tex}
\input{7Conclusion.tex}

\textbf{Acknowledgements.} This work has been funded by Deutsche Forschungsgemeinschaft within the project \textit{ExpLAIN. Between the Lines -- Knowledge-based Analysis of Argumentation in a Formal Argumentation Inference System}, FR 1707/-4-1, as part of the RATIO Priority Program.
We also want to thank our reviewers, our annotators Angel Daza and Esther van den Berg, and NVIDIA for donating GPUs.

\bibliography{naaclhlt2019}
\bibliographystyle{acl_natbib}

\end{document}

%% file: 1Intro.tex
\section{Introduction}
\label{sec:intro}
Commonsense knowledge can be seen as a large amount of diverse but simple facts about the world, people and everyday life, e.g., \textit{Cars are used to travel} or \textit{Birds can fly} \citep{liebermann2008}. Commonsense knowledge obtained from \cn~is increasingly used in advanced NLU tasks, such as  textual entailment \citep{weissenborn2018}, reading comprehension \citep{mihaylovfrank:2018}, machine comprehension \citep{Wang2018, Gonzalez18}, question answering \citep{semeval18-11} or dialogue modeling \citep{Young2018Augmenting} and also applications in vision \citep{Le13}.
Some of these approaches exploit embeddings learned from \cn, others select specific relations from it, depending on the application. 

This paper proposes a multi-label neural approach for classifying \cn~relations, where the task is to predict one (or several) commonsense relations from a given set of relation types that hold between two given concepts from \cn. In future work, the predicted relations can then be used for enriching \cn~by adding relations between concepts which are not yet linked in the network.

We design the task of multi-label neural relational classification to account for specific properties of \cn:

(i) \cn's relation inventory is not designed to be disjunct: a given pair of  relation arguments (in \cn: \textit{concepts}) may be connected by 
more than one relation type: e.g.\\ 
 \tr{people}{Desires/CapableOf}{eating in groups},  
\tr{reading}{UsedFor/Causes}{education}. 
This places relations in close vicinity in semantic space, making  relation prediction a hard task. 

(ii) Concepts often are \textit{multi-word expressions} of \textit{different phrase types} (e.g., noun or verb phrases),
posing a challenge for argument representation.
Relation slots may also be filled by \textit{different semantic types}: e.g., the 2nd argument of \textsc{desires} can be an entity or event. Such heterogeneous signatures increase classification difficulty.

(iii) As any knowledge resource, \cn~is incomplete, which means that relations between concepts are missing.
The incompleteness of the resource poses serious evaluation problems, since assumed negative instances may in fact be positive. 

To tackle these issues we perform a thorough experimental examination of the learnability of \cn~relations in a controlled
multi-label classification setting. 
Our contributions are:
(i) a cleaned and balanced data subset covering the 14 most frequent relation types from the core part of \cn~that serves as a basis for 
assessing
relation-specific  classification performance.
We extend this dataset to an open-world classification setup;
(ii) a neural multi-label classification approach with various model options for the representation of relations and their (multi-word) arguments, including relation-specific label prediction thresholds;
(iii) an in-depth analysis of specific properties of the \cn~relation inventory, from which we derive hypotheses that we evaluate in  classification experiments;
(iv) we perform detailed analysis of results that  confirm a great number of our hypotheses regarding specific classification challenges; (v) finally,  we assess the amount of potential evaluation discrepancies due to the incompleteness of the resource 
in a small-scale annotation experiment. 




%% file: 2RelWork.tex
\section{Related Work}
\subsection{Semantic Relation Classification} 
Semantic relation classification covers a wide range of methods and learning paradigms for representing relation instances (see \citeauthor{nastase2013semantic} \citeyear{nastase2013semantic} for an overview). Typically, the data is presented to the learner as independent instances, with
or without
a sentential context. 
Relation classification models 
represent
the meaning of the arguments (attributional features) and if context is available, also the relation (relational features). 

Recently Deep Learning has strongly influenced semantic relation learning. Word embeddings can provide attributional features for a variety of learning frameworks \citep{attia2016, vylomova-EtAl:2016:P16-1},
and the sentential context -- in its entirety, or only the structured (through grammatical relations) or unstructured phrase expressing the relation -- can be modeled through a variety of neural architectures -- CNN \citep{zhen18, Ren18} or RNN variations \citep{zhang18}. 

\subsection{\cn~Relation Classification}
\cite{Speer:2008:ARD:1619995.1620084} introduce AnalogySpace, a representation of concepts and relations in \cn~built by factorizing a matrix with concepts
on one axis and their features or properties (according to \cn) on the other. This low-dimensional representation allows for finding analogous facts, generalizations, new categories and justifications for 
classifications based on known properties. While this representation allows for recomputing the confidence of existing facts, the focus 
was not on classifying or trying to learn specific relations represented in the resource.

\citet{li-EtAl:2016:P16-14} apply \textit{matrix factorization} to \cn~with the aim of resource extension and report 
91\% accuracy in a \textit{binary} evaluation (i.e., verifying the correctness of an  (unlabeled) link between concepts).
\citet{saito18} expand this work by combining the knowledge base completion task (distinguishing true relation triples consisting of arbitrary phrases from false ones) with the task of knowledge generation (finding the second entity for a given first entity and a given relation). They enhance the link prediction model of \citeauthor{li-EtAl:2016:P16-14}\ with a 
model that 
learns the two tasks -- knowledge base completion and knowledge generation -- 
jointly and 
outperform the completion accuracy results of \citeauthor{li-EtAl:2016:P16-14} 
by up to 3pp.

Many NLU tasks rely on \textit{specific relations} from \cn~\citep{Le13,Shudo16}. It is thus important to  assess classification accuracy for individual relation types. 

%% file: 4RelClass.tex
\section{
The Difficulty of \cn~Relation Classification}


\subsection{\cn~
Dataset}\label{sec:data}

The Open Mind Common Sense (\textsc{Omcs}) project \citep{Speer:2008:ARD:1619995.1620084} started the acquisition of common sense knowledge from contributions over the web, leading to \cn, which now also includes expert-created resources (such as WordNet) and automatically extracted knowledge or knowledge obtained through games with a purpose \citep{Speer:2008:ARD:1619995.1620084}. 
The current version, \cn~5.6, comprises 37 relations, some of which are commonly used in other resources like WordNet (e.g.\ \rel{IsA, PartOf}) while most others are more specific to capturing commonsense information and 
as such 
are 
particular to \cn~(e.g.\ \rel{HasPrerequisite}, 
\rel{MotivatedByGoal}). With very few exceptions (e.g., \rel{Synonym} or \rel{Antonym}), \cn-relations are asymmetric. 
The English version consists of 1.9 million concepts and 1.1 million links to other databases, such as DBpedia. In our work we focus on the English \textsc{Omcs} subpart  (\textsc{Cn-Omcs}). 

\subsection{Task Definition}

Given a pair of concepts $\langle c_i, c_j\rangle$, where $c_i, c_j$ may be multi-word expressions, the task is to automatically predict one (or several, see \textsection \ref{sec:multimodel} for the multi-label aspect of the task) commonsense relations $r_t$ from a given set of \cn~relation types $R_{\textsc{cn}}$ that hold between c$_i$ and c$_j$. Relations are presented to the classifier without textual context, and thus a crucial aspect is using a representation that properly captures the semantics of the arguments.

\subsection{Designing a Relation Classification System for \cn}
\label{sec:hypotheses}

\cn~has very specific properties in terms of the relations included, the type of the arguments, coverage and completeness. A successful relation classification system should take these into account.
Given the heterogeneity of sources of \cn, we focus on its core part, in particular \textsc{Cn-Omcs-Cln}, a subset selected from \textsc{Cn-Omcs} that includes ca.\ 180K triples from 36 relation types, restricted to known vocabulary from the GoogleNews Corpus (see \textsection \ref{sec:data} for further details).

\subsubsection{Representing the Inputs}\label{sec:repr}

Word embeddings have been shown to provide useful semantic representations, capturing 
lexical properties of words and relative positioning in 
semantic space  
\citep{linguistic-regularities-in-continuous-space-word-representations}, which has been exploited for semantic relation classification \citep{vylomova-EtAl:2016:P16-1,attia2016}. 

Following this work, we represent a \textit{pair of concepts} $\langle c_i, c_j\rangle$ whose relation we want to classify through their embeddings  $v_{c_{i}}$ and $v_{c_{j}}$. These argument representations can be combined 
by  subtraction 
$(v\textsubscript{c\textsubscript{i}}- v\textsubscript{c\textsubscript{j}})$
(\textit{DiffVec}; cf. \citeauthor{weeds14} \citeyear{weeds14}, \citeauthor{roller14} \citeyear{roller14}),  addition 
$v\textsubscript{c\textsubscript{i}}+ v\textsubscript{c\textsubscript{j}}$
(\textit{AddVec})
or  concatenation
$[v\textsubscript{c\textsubscript{i}},v\textsubscript{c\textsubscript{j}}]$ (\textit{ConcatVec}, cf. \citeauthor{baroni-etal-2012-entailment} \citeyear{baroni-etal-2012-entailment}).

One of the issues in using such representations for \cn~is the fact that most \cn~concepts are multi-word expressions (1.93 words on average, cf.\ Table \ref{tbl:relations_statistics}). 
We experiment with two ways of producing a representation for a multi-word concept: (i) computing a \textit{centroid vector}, as the normalized 
sum over the embedding vectors of all words in the expression (as the baseline); (ii) encoding the expression using an RNN, e.g.  
a (Bi)LSTM, which encodes sequences of various lengths into one fixed-length vector.
We hypothesize that 
using an RNN
yields better concept representations than 
centroid vectors.

\begin{figure}[t]
\centering
\includegraphics[scale=0.5]{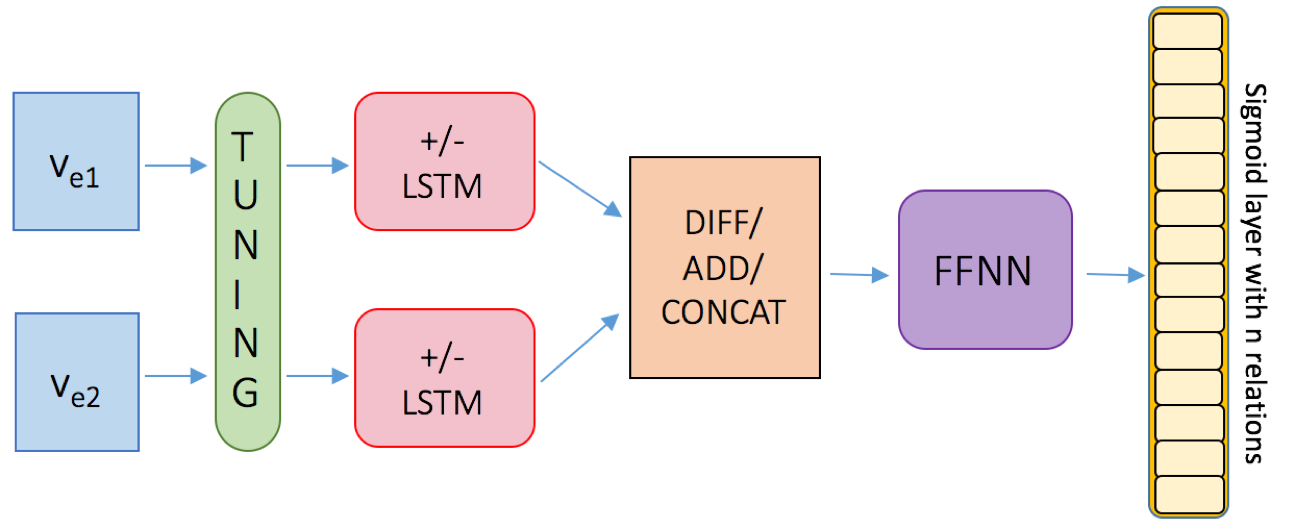}
\caption{Multi-Label Classification Model. 
} 
\label{fig:model}
\end{figure}

\subsubsection{Constructing a Multi-Label Classifier}
\label{sec:multimodel}

\input{tables/freq_relations.tex}

An important characteristic of \cn~is that more than one relation can hold for a given pair of concepts. On average this applies to 5.37\% of instances per relation (cf.\  \ref{tbl:relations_statistics}).
Consequently, we cast our classification task as a \textit{multi-label classification problem}. 
\input{tables/relation_set_new_onlycn.tex}

\paragraph{Model architecture.}
Fig.\ \ref{fig:model} illustrates the model architecture. 
Input concept pairs are encoded -- as centroids or using RNNs -- and the representations are combined and presented to a feed-forward neural network (\textsc{Ffnn}) with one hidden layer to  non-linearly separate the relation classes.

In single-label classification, the probability for a class is not independent from the other class probabilities. Hence, $softmax$  is typically used at the output layer. By contrast, in multi-label classification, we want to model class predictions individually.
The $sigmoid$ 
models the probability of a label as an independent \textit{Bernoulli} distribution:
\begin{equation*}
\begin{split}\begin{array}{@{}ll@{}}
sig(t) = \frac{1}{1+e^{-t}} = \frac{1}{2}(1+tanh \frac{t}{2})
\end{array}\end{split}
\end{equation*}
This actually translates to an independent binary neural network for each label, resulting in a set of isolated binary classification tasks (cf. \citeauthor{Sterbak17} \citeyear{Sterbak17}; \citeauthor{He2018} \citeyear{He2018}). 

The \textsc{FFnn} uses $sigmoid(\sigma(xW^h)W^o)$,
where $x$ is the input vector and $W^h$ and $W^o$ are weight matrices. 
We use binary cross entropy as our loss function. The architecture allows us to tune pre-trained embeddings for our relation learning task. 
\\The hypotheses arising from the multi-label setting of \cn~are: (i) discrimination of overlapping classes is more difficult, compared to the usual relation classification task with disjoint relations (e.g.\  \citeauthor{hendrickx-EtAl:2010:SemEval}\ \citeyear{hendrickx-EtAl:2010:SemEval}).
(ii) given the incompleteness of \cn, the classification performance may be
erroneously assessed due to missing relations in the data. We will estimate the effect of this phenomenon in a  small-scale annotation experiment.

\subsubsection{
Relation Classification Difficulty and Relation-specific Thresholding} 
\cn~relation types show  great divergence 
with respect to their argument's semantic and phrase types,
as shown in  \ref{tbl:relations_statistics}. About half of the relation arguments are nominal, 
entity-denoting concepts, with location as a specific entity
type, half of them are event-type arguments. 
Several relations take different semantic types in a single argument position (e.g., \rel{HasSubevent, Causes}). Diversity of semantic types and phrase types -- especially within 
a single argument position
-- is a challenge for relation learning. 
We expect 
classification to be more difficult on relations with a mixture of argument types. 
Because of this, different thresholds may be needed for predicting different relation types. We adopt
a customized multi-label prediction
setup where we tune thresholds separately for each relation type.
We expect that individually tuned,  relation-specific
thresholds improve overall classification performance. 

%% file: tables/freq_relations.tex
\begin{table} [b]
\centering
\begin{tabular}{ll|ll|ll|ll}
\hline
\rel{UsedFor}&33290&\rel{HasPrereq.}&16565&\rel{HasProp.}&5782&\rel{Has1Sub.}&2697\\
\rel{AtLocation}&23874&\rel{IsA}&15063&\rel{Rec.Action}&4494&\rel{Desires}&2584\\
\rel{HasSubevent}&20518&\rel{Causes}&12414&\rel{HasA}&4118\\
\rel{CapableOf}&18909&\rel{Mot.ByGoal}&7243&\rel{CausesDes.}&3587\\
\hline

\end{tabular}
\caption{Number of instances per 14 most frequent relations in \textsc{Cn-Omcs-Cln}.
}
\label{tbl:freq}
\end{table}

%% file: tables/relation_set_new_onlycn.tex
\begin{table*}[t]
\centering{
\small{
\scalebox{0.99}{
\centering
\begin{tabular}{@{}lllll|ccc}
\toprule
 &\multicolumn{2}{c}{semantic type}& \multicolumn{2}{c}{phrase type}& multiword & words per & multi-label \\

relations & \textsc{arg$_1$} & \textsc{arg$_2$}& \textsc{arg$_1$} & \textsc{arg$_2$}&concepts (\%) & concept (avg) & relations (\%) \\ 
\midrule
\textsc{IsA} & \cellcolor{red!40}entity & \cellcolor{red!40}entity& NP & NP&46.01 &1.78 &1.17 
\\
\textsc{HasA} & \cellcolor{red!40}entity & \cellcolor{red!40}entity & NP & NP&56.36 &1.96 &1.07 
\\
\textsc{AtLocation} & \cellcolor{red!40}entity&\cellcolor{gray!25}location & NP & NP& 34.66 &1.42 &1.06 
\\
\textsc{HasProperty}& \cellcolor{red!40}entity& \cellcolor{cyan!25}property& NP & AP&41.41 &1.80 &1.31 
 \\
\textsc{UsedFor} & \cellcolor{red!40}entity & \cellcolor{yellow!35}event& NP & VP&63.98 &1.95 &4.41 
\\
\textsc{CapableOf}& \cellcolor{red!40}entity & \cellcolor{yellow!35}event& NP &VP&58.21 &1.89 &1.17 
 \\
 \textsc{ReceivesAction}& \cellcolor{red!40}entity& \cellcolor{yellow!35}event& NP & VP&57.78 &2.24 &0.18 
 \\
\textsc{CausesDesire} & \cellcolor{red!40}entity& \cellcolor{yellow!35}event& NP & VP&74.35 &2.10 &0.67 
\\
\textsc{Desires} & \cellcolor{red!40}entity & \cellcolor{yellow!35}ev/\cellcolor{orange!70}entity& VP &V/NP&40.02 &1.68 &2.79 
\\
 \textsc{Motiv.ByGoal} & \cellcolor{yellow!35}event & \cellcolor{yellow!35}event &VP&VP&79.33 &2.21 &6.57 
\\
\textsc{HasPrerequisite}& \cellcolor{yellow!35}event & \cellcolor{yellow!35}event&VP&VP&80.27 &2.23 &9.86 
 \\
\textsc{HasFirstSubev.} & \cellcolor{yellow!35}event & \cellcolor{orange!70}ev/entity& VP &V/NP&83.89 &2.27 &36.26 
\\
\textsc{HasSubevent}& \cellcolor{yellow!35}event & \cellcolor{orange!70}ev/entity& VP &V/NP&80.21 &2.21 &11.02 
\\
\textsc{Causes} &\cellcolor{orange!70}ev/entity & \cellcolor{orange!70}ev/entity&V/NP& V/NP&73.42 &2.10 &12.47 
\\
\hline

All relations  &  &  &  &  & 61.01 &1.93 &5.37 
\\
\bottomrule
\end{tabular}
}}
\caption{14 most frequent relations in \textsc{Cn-Omcs-Cln}: their semantic and phrase types (col.\ 2-5); percentage of multiword concepts  (col.\ 6); average number of words per concept (col.\ 7); percentage of relation instances 
for which we find another relation instance in \textsc{Cn-Omcs-Cln} 
(col.\ 8).  
}
\label{tbl:relations_statistics}
}
\vspace{-0.1in}
\end{table*}

%% file: 5Experiments.tex
\section{Experiments}
\label{sec:experiments}

\subsection{Dataset Construction}

\paragraph{\textsc{Cn-Omcs-cln}}
\label{sec:data}

\textsc{Cn-Omcs} contains noise in form of typos, unknown words, or words from other languages than English.
\footnote{We find 
a lot of Chinese words with the English tag \textit{en} in \textsc{cn-Omcs}.}
We check all relation triples in \textsc{Cn-Omcs} against the vocabulary of 
\textit{word2vec} embeddings trained on part of the Google News dataset.\footnote{About 100 billion words, cf. \citeauthor{Mikolov13}  \citeyear{Mikolov13}.}
This embedding set contains vectors for 3 million words and phrases. 
We discard all relation triples from \textsc{Cn-Omcs} which contain words that do not appear in this set. 
The resulting dataset -- \textsc{Cn-Omcs-cln} -- contains 179.693 triples drawn from 36 different relations (relation distribution displayed in Table \ref{tbl:freq}). The \rel{Other} class comprises all relations from \textsc{Cn-Omcs-cln} 
with
less than 2000 instances.\\

\textbf{\textsc{Cn-Omcs-14}} Based on \textsc{Cn-Omcs-Cln}  we construct our experimental dataset \textsc{Cn-Omcs-14} -- a balanced dataset 
still large enough 
for
applying
neural methods.
We 
include all relations from \textsc{Cn-Omcs-cln} with more than 2000 instances, and downsample to the least frequent class -- 2586 instances per relation.
 To select the \say{best} instances for testing and tuning, we
sort the relation triples by their confidence score, as provided by \cn.
Inspired by \citet{li-EtAl:2016:P16-14} we select the 10\% (258) most confident tuples per relation for testing, the next 10\% for development, the remaining 80\% (2068) for training, cf.\ Table \ref{tbl:datasets}.


\textbf{Closed vs.\
Open World Setting.}
Learning to classify relations in a closed world setting is limited to the relation types present in the data. We want to design
a system that is also able to detect whether \textit{a relation exists} between concepts -- but none of the provided ones, or whether 
\textit{no 
relation} holds. We thus
extend
the data set 
with two classes: \textbf{\rel{Other}} -- containing concept pairs that \textit{do stand} in a relation, yet not any of those present 
in the target relation set; and \textbf{\rel{Random}} -- containing concept pairs that are \textit{not related}.

Instances for the \rel{Other} class consist of a sample of triples from the 22 low-frequency relations that were not included in \textsc{Cn-Omcs-14}, these are the following relations: \rel{MadeOf}, \rel{DBPedia}, \rel{RelatedTo}, \rel{Difference}, \rel{LocatedNear}, \rel{CreatedBy}, \rel{NotUsedFor}, \rel{FormOf}, \rel{DerivedFrom}, \rel{ObstructedBy}, \rel{Synonym}, \rel{PartOf}, \rel{SymbolOf}, \rel{NotDesires}, \rel{HasContext}, \rel{DefinedAs}, \rel{HasLastSubevent}, \rel{ExternalURL}, \rel{InstanceOf}, \rel{NotCapableOf}, and \rel{NotHasProperty}.  

\input{tables/datasets.tex}

Instances for the \rel{Random} class are generated similarly to 
\citet{vylomova-EtAl:2016:P16-1}:  50\% of instances are \textit{opposite pairs}, obtained by switching the order of concept pairs within the same relation;  
 50\% are  \textit{corrupt pairs}, obtained by replacing one concept in a connected pair with a random concept from the same relation.
Using
\textit{corrupt pairs} ensures that our model does not simply learn  properties of the word classes, but instead is forced to
encode relation instances. \rel{Random} and \rel{Other} are the same size as the individual target relations.




\subsection{Experiment Setup}

\textbf{Experiments and Datasets.} 
 We experiment with two open world settings: in OW-1 we add only the \rel{Random} class to \textsc{Cn-Omcs-14}, to investigate whether the classifier is able to differentiate related from non-related concept pairs.
in  OW-2 we add both 
\rel{Other} and \rel{Random} to \textsc{Cn-Omcs-14}, to investigate whether the classifier can also learn to predict that an unknown relation exists or that no relation holds.
We also report results of the closed world setting where we exclude \rel{Other} and \rel{Random}.
Each dataset is split into training (80\%), dev (10\%) and test (10\%)  (cf. Table \ref{tbl:datasets}).\\
\textbf{Evaluation.} 
We evaluate model performance in terms of F1 score for each relation. 
We report averaged weighted F1 scores over 5 runs.

\subsection{Model Parameters} 

\paragraph{Embeddings.}
\label{sec:emb}

Based on preliminary experiments\footnote{We additionally tested Numberbatch embeddings \citep{speer17}, GloVe embeddings trained on Wikipedia and Gigaword \citep{pennington2014glove}, context2vec embeddings trained on UkWaC \citep{MelamudGD16}. In our experiments we discovered that all of these alternatives perform worse than the \textit{word2vec} embeddings.}, we use 300-dim.\ skip-gram
\textit{word2vec} embeddings trained on part of the Google News dataset (100 billion words, \citeauthor{Mikolov13} \citeyear{Mikolov13}). 
Embeddings are tuned
during training.\\
\textbf{Concept representation.} Concept are encoded using centroid vectors or an RNN (cf.\ \textsection \ref{sec:repr}).\\
\textbf{Relation representation.}
We use the \textit{ConcatVec} representation (\textsection \ref{sec:repr}), which we determined to be the most useful in preliminary experiments.\\
\textbf{Label prediction thresholds}
are tuned in two ways: (i) a global threshold for all relations and (ii) separately tuned thresholds for each relation.\\
\textbf{Hyperparameter settings} were determined on the devset. 
For encoding of multiword terms we use bi-LSTMs with one hidden layer and a cell size of 350 (perform better than GRUs and LSTMs).
For the \textsc{Ffnn} we tune the hidden layer size and the activation function. Optimal hyperparameters are 200 (\textsc{Ffnn}), 100 (\textsc{Ffnn+Rnn}), and $ReLU$ for both \textsc{Ffnn} and \textsc{Ffnn+Rnn}.\\
\textbf{Implementation.} We implemented our models with \textit{PyTorch} \citep{pytorch}. 

%% file: tables/datasets.tex
\begin{table}[t]
\centering
\scalebox{1.1}{
\small{
\begin{tabular}{@{}l|cccc}
\toprule
dataset & relations & train & dev & test \\
\midrule
CN-OMCS-14 CW                 & 14  & 28,952 & 3612 &3612\\
CN-OMCS-14 OW-1                & 14+1  & 30,960 & 3870 & 3870\\
CN-OMCS-14 OW-2                & 14+2  & 33,088 & 4128 & 4128\\

\bottomrule
\end{tabular}
}}
\caption{Dataset details and splits.
}
\label{tbl:datasets}
\vspace{-0.1in}
\end{table}

%% file: 5results.tex
\subsection{Results}
\label{sec:results}

Table \ref{tbl:results} summarizes the results
in open (OW) and closed world (CW) settings. 

The overall best performing model across all settings is \textsc{FFnn+Rnn} (as opposed to \textsc{FFnn} with centroid argument representations) with relation-specific label prediction thresholds (as opposed to one global threshold value). In the OW setting we achieve overall F1-scores of 0.68 
(OW-1) and 0.65 
(OW-2). The CW setting leads to best results with 0.71 F1. 
The models improve by 4pp (OW-1), 7pp (OW-2) and 6pp (CW) when replacing centroids with bi-LSTM encoded concept representations.
Relation-specific thresholds improve results by 2pp (FFNN+RNN on OW-1, OW-2 and CW).
Across all settings (CW, OW-1, OW-2) the best performing relations 
are: 
\rel{Desires} (0.94), \rel{ReceivesAction} (0.91), \rel{CausesDesire} (0.90). 
We observe lowest F1-scores for \rel{HasSubevent} (0.26, 0.24), \rel{HasPrerequisite} (0.38, 0.39) and \rel{HasFirstSubevent} (0.55, 0.61) in OW-1 and CW, respectively.
The \textsc{random} and \rel{other} classes have poor results overall. OW-2 with two OW classes performs worse than OW-1 with the single \rel{random} class.
The low results on the \rel{other} class (0.40) could stem from its heterogeneity. The system finds it difficult to differentiate \rel{other} and \rel{random}.

\input{tables/results_new.tex}



%% file: tables/results_new.tex
\begin{table}[t]
\centering
\scalebox{.75}{
\begin{tabular}{l@{\hspace*{12mm}}c@{\hspace*{12mm}}c@{\hspace*{19mm}}c@{\hspace*{12mm}}c@{\hspace*{19mm}}c@{\hspace*{12mm}}c@{\hspace*{12mm}}c@{\hspace*{1mm}}c@{}}
\toprule
Setting &  \multicolumn{2}{c}{OpenWorld OW-1} &  \multicolumn{2}{c}{OpenWorld OW-2} &\multicolumn{2}{c}{Closed World}  \\
\hline
Model &  FF & FF\small{+RNN}&  FF & FF\small{+RNN}&  FF & FF\small{+RNN}\\
\hline \hline
\small{\rel{IsA}}& .58 (.57) & .62 (.60) & .51 (.51) & .60 (.57) & .64 (.63) & .67 (.67)\\
\small{\rel{HasA}} &.67 (.66) & .80 (.79) & .53 (.52) & .79 (.77)& .73 (.72) & .80 (.78)\\ 
\small{\rel{AtLocation}} & .69 (.68) & .78 (.78) & .63 (.61) & .74 (.72) & .77 (.75) & .84 (.83)\\
\small{\rel{HasProperty}} & .66 (.65) & .81 (.80) & .62 (.61) & .78 (.77)& .67 (.67) & .84 (.83)\\
\small{\rel{UsedFor}} & .76 (.75) & .78 (.77) &  .79 (.78) & .76 (.76) & .78 (.78) & .79 (.78)\\
\small{\rel{CapableOf}} & .61 (.61) & .67 (.65) & .56 (.56) & .65 (.64) & .61 (.60) & .71 (.71)\\
\small{\rel{ReceivesAction}} & .82 (.82) & .91 (.91) & .77 (.77) & .90 (.90)& .87 (.86) & .93 (.93)\\   \small{\rel{Caus.Des.}} &.87 (.87) & .90 (.88) & .86 (.85) & .87 (.87)& .87 (.86) & .92 (.90)\\
\small{\rel{Desires}} &.91 (.85) & .94 (.92) &  .73 (.65) & .93 (.88)& .87 (.83) & .88 (.94)\\ 
\small{\rel{MoticatedByGoal}} & .61 (.60) & .56 (.55) & .56 (.55) & .59 (.59)& .60 (.59) & .64 (.61)\\
\small{\rel{HasPrerequisite}}& .45 (.41) & .38 (.36) & .38 (.36) & .38 (.36) & .43 (.42) & .39 (.38) \\
\small{\rel{HasFirstSubevent}} &.54 (.53) & .55 (.55) & .49 (.49) & .56 (.55)& .51 (.50) & .61 (.60)\\
\small{\rel{HasSubevent}} & .24 (.22) & .26 (.16) & .17 (.15) & .24 (.15) & .21 (.21) & .24 (.20)\\
\small{\rel{Causes}}& .60 (.59) & .57 (.56) & .59 (.58) & .61 (.60) & .61 (.60) & .61 (.61)\\\hline

\small{\rel{Other}} & - & - & .39 (.39) & .40 (.40) & - & -\\
\small{\rel{Random}} & .61 (.58) & .59 (.54) & .62 (.61) & .53 (.49)& - & -\\ 
\hline \hline
Weighted F1 & .64 (.63) & .68 (.66) & .58 (.56) & .65 (.63) & .65 (.64) & .71 (.69) \\ \hline

\bottomrule
\end{tabular}
}
\caption{Weighted F1 results on \textsc{Cn-Omcs-14}. 
Main results obtained with relation-specific prediction thresholds (in brackets: results for global prediction threshold).} 
\label{tbl:results}
\end{table}

%% file: 6Analysis.tex
\section{Analysis}
\label{sec:analysis}

\input{tables/relation_set_cn2000.tex}

\begin{figure*}[b!]
\centering
\includegraphics[scale=0.9]{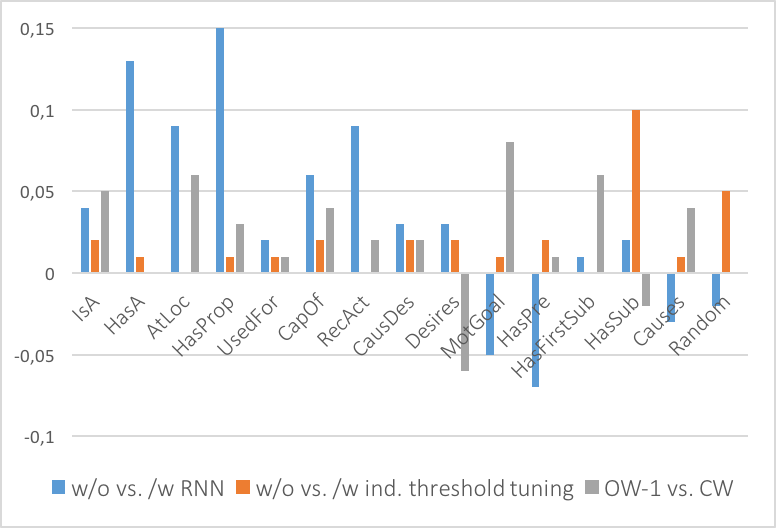}
\caption{Delta of F1-scores on \textsc{Cn-Omcs-14}: (i) \textsc{FFnn} vs.\  \textsc{FFnn+Rnn}: (relation-specific\ threshold, OW-1, 
blue); (ii) global vs.\ relation-specific~threshold\ (\textsc{FFnn+Rnn},OW-1, 
orange); (iii) OW-1 vs.~CW (\textsc{FFnn+RNN}, relation-specific~threshold, 
grey).} 
\label{fig:delta}
\end{figure*}
\begin{figure*}[b!]
\centering
\includegraphics[scale=0.5]{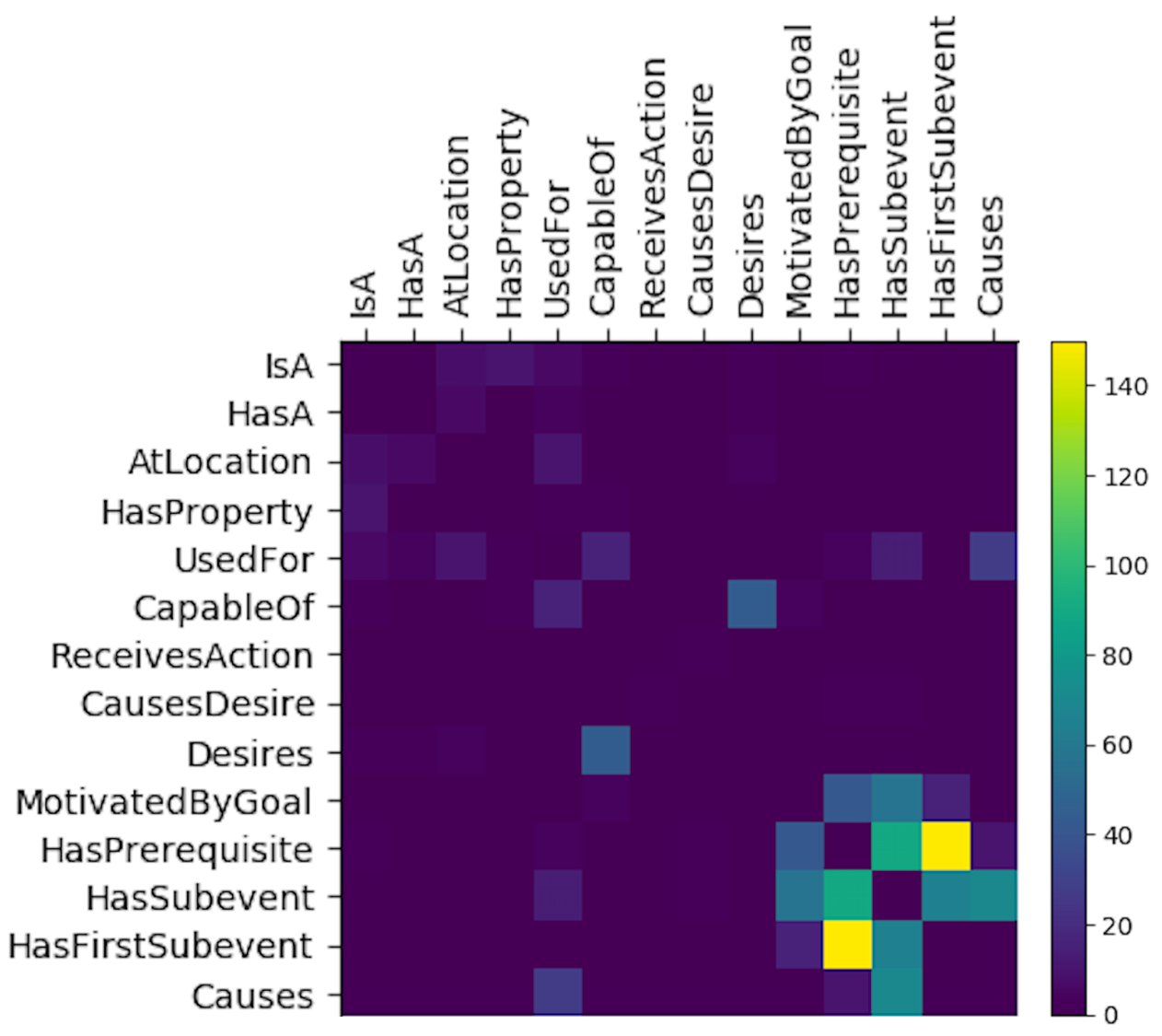}
\caption{Visualizing ambiguous (co-occurring) relations in 
\textsc{Cn-Omcs-14} in a  symmetric heatmap.} 
\label{fig:heatmap}

\label{fig:x}
\end{figure*}

In this section we will discuss the hypotheses derived from our analysis of \cn~properties (\textsection \ref{sec:hypotheses}), and based on that, to determine which approaches and representations are best suited for \cn-based commonsense relation classification. To aid the discussion we produced Figures \ref{fig:delta}, \ref{fig:heatmap}, \ref{fig:interrelations}, and Table \ref{tbl:statistics_cn2000}.  \\
Fig.\ \ref{fig:delta} plots differences in performance for each relation for the setting we wish to compare: \textit{concept encoding} using centroids (\textsc{FFnn}) vs.\ RNNs (\textsc{FFnn+Rnn}) (
blue), \textit{global vs.\ relation-specific} prediction threshold (
orange), and OW-1 vs. CW setting (
grey). \\
Fig.\ \ref{fig:heatmap} visualizes  ambiguous -- that means  co-occurring --  relations  in our dataset in a symmetric heatmap. \\
Fig.\ \ref{fig:interrelations} displays interrelations between concept characteristics and model performance,
based on our best performing system (FFNN+RNN+ind. tuned thresholds, OW-1). To observe correlations between classification performance and different measurable characteristics of the data in Fig.\ \ref{fig:interrelations}, we  scaled the following values for each relation to a common range of 0 to 15: the percentage of multi-word terms 
(cf.\ Table \ref{tbl:relations_statistics}) (
grey), the average number of words per concept (cf.\ Table \ref{tbl:relations_statistics}) (
yellow), percentage of relation instances with multiple labels (cf.\ Table \ref{tbl:relations_statistics}) (
blue), best model performance on OW-1 (\textsc{FFnn+Rnn} with individually tuned thresholds, cf.\ Table \ref{tbl:results}) (
red) and the corresponding relation-specific thresholds
(
green).\\
Table \ref{tbl:statistics_cn2000} gives relation statistics on \textsc{Cn-Omcs-14} (as opposed to Table \ref{tbl:relations_statistics}, which gives statistics for the complete version \textsc{Cn-Omcs-Cln}).

~~~

\subsection{
Representing Multi-word Concepts
} 
\label{sec:mwe}



\begin{figure}[t]
\centering
\includegraphics[scale=0.8]{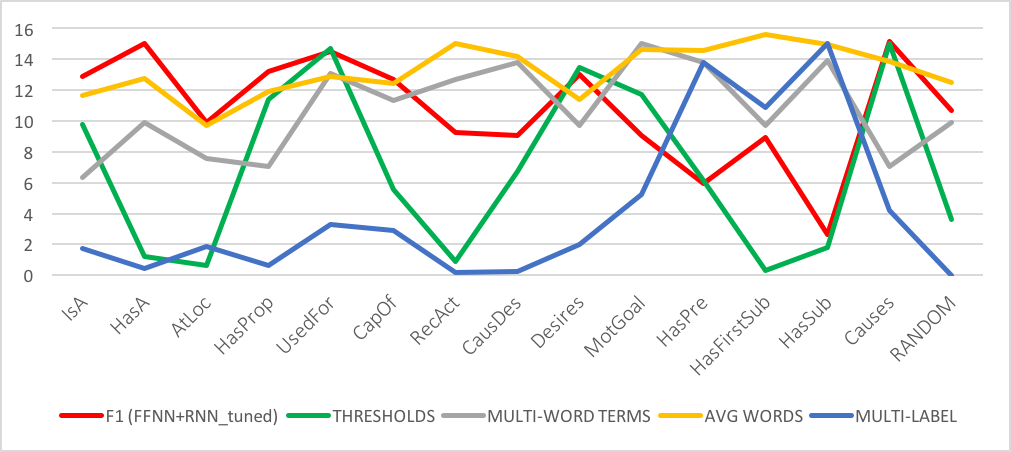}
\caption{
Interrelations between concept characteristics and model performance, 
based on 
FFNN+RNN+individually tuned thresholds, OW-1, 
scaled to range 0 to 15.} 
\label{fig:interrelations}
\end{figure}

We hypothesized that there is a correlation between the length of the arguments and 
model performance when encoding arguments with an RNN. We find no such correlation -- the relations that benefit the most from using an RNN (Fig.\ \ref{fig:delta}: blue and Fig.\ \ref{fig:interrelations}: yellow, red) are not those with the longest arguments (cf. Table \ref{tbl:statistics_cn2000}). Instead we find that the relations \rel{HasProperty}, \rel{HasA}, \rel{AtLocation}, and \rel{ReceivesAction} benefit most from concept encoding with a RNN, followed by \rel{CapableOf, IsA, Desires, CausesDesire} and \rel{Has(First)Subevent} with lower margins. The missing correlation can be confirmed by a very low Pearson's coefficient of only 0.05 between (1) improvements we get from enhancing \textsc{FFnn} with RNN (i.e., delta of F1 scores for \textsc{FFnn} vs.\ \textsc{FFnn+Rnn}; both with individually tuned thresholds) and (2) the average number of words per concepts (cf. Table \ref{tbl:relations_statistics}). 

\subsection{Threshold Tuning \& Model Performance}

We hypothesized that relations would benefit from having individually tuned thresholds. Overall, the models with RNN encoding of concepts
benefit more from threshold tuning than the basic \textsc{FFnn}. 
Regarding single relations (Fig.\ \ref{fig:delta}, orange bars),  \rel{HasSubevent} and the open world class \rel{Random} benefit the most from individual threshold tuning (both with relatively low F1 scores).
The individual thresholds vary considerably across relations (Fig.\ \ref{fig:interrelations}).

To test whether relations that are harder to classify benefit the most from tuning the threshold (as the performance of \rel{HasSubevent} and \rel{Random} seem to indicate), we compute the correlation between (1) the difference of model performance with and without individually tuned thresholds (as described above) and (2) general model performance (F1 scores of FFNN+RNN with global threshold, OW-1). The score of -0.67 Pearson correlation indicates that indeed relations with lower general performance will tend to have higher improvements.
This is also reflected in Fig.\ \ref{fig:interrelations} (green and red), which also shows that for relations with higher F1 scores, higher thresholds tend to work better.
Relation classification models applied to \cn~should therefore have higher thresholds for relations with high classification confidence (high F1 scores), while for relations with low performance lower thresholds are recommended.

\subsection{Closed vs.\ Open World  Setting}

Most relations perform better in the CW setting (cf.\ grey bars in Fig.\ \ref{fig:delta}), especially \rel{MotivatedbyGoal}, \rel{HasFirstSubevent}, \rel{AtLocation}, and \rel{IsA} (Fig.\ \ref{fig:delta}, grey). In contrast, \rel{Desires} and \rel{HasSubEvent} perform better in an open world setting (Fig.\ \ref{fig:delta}). Comparing the two settings OW-1 and OW-2 (Table \ref{tbl:results}, not displayed in Fig.\ \ref{fig:delta}), we find that only the relations \rel{MotivatedBy, HasFirstSubevent} and \rel{Causes} perform better in OW-2 than in OW-1. All other relations benefit from the OW-1 setting, especially \rel{AtLocation} and the open world class \rel{Random}.


\subsection{Relation Heterogeneity}
We hypothesized that relations that are more heterogeneous with respect to the type of their arguments (whether semantic or phrasal) will be harder to learn.
Comparing the degree of diversity of semantic or phrase types (Table \ref{tbl:relations_statistics}) with model performance confirms this hypothesis. The relations that perform best have semantically or \say{phrasally} consistent arguments,
whereas (apart from \rel{Desire}) relation types that feature different types of entities or phrases in the same argument position tend to achieve low F1 scores.

\subsection{Relation Ambiguity}

We hypothesized that relations that have multi-labeled instances (instances to which more than one label -- relation -- applies) will be more difficult to learn.
Fig.\ \ref{fig:heatmap} illustrates relation co-occurrences, i.e.\  relations that have overlapping instances.
The most frequently co-occuring relations in \textsc{Cn-Omcs-14} are \textsc{HasPrerequisite} \& \textsc{HasFirstSubevent}, (150 co-occurrences), \textsc{HasSubevent} \& \textsc{HasPrerequisite} (90) and \textsc{HasSubevent} \& \textsc{Other} (86).\footnote{In \textsc{Cn-Omcs-Cln} (complete, unbalanced dataset) the most frequently overlapping relations are: \textsc{UsedFor} \& \textsc{Causes} (800), \textsc{HasSubevent} \& \textsc{Causes} (636), and \textsc{HasSubevent} \& \textsc{HasPrerequisite} (628).}
399 concept pairs have two relation labels (e.g., $\langle$\textit{a cat,meow}$\rangle$: \textsc{Desires,  CapableOf}), 20 pairs have three (e.g., $\langle$\textit{playing a harp,making music}$\rangle$: \textsc{Causes,	HasSubevent, UsedFor}), and two pairs have four: $\langle$\textit{opening a gift,surprise}$\rangle$: \textsc{HasPrerequisite, Causes, HasSubevent, UsedFor}.

Fig.\ \ref{fig:interrelations} shows 
a strong inverse correlation (-0.82 Pearson) between model performance 
and the number of multi-labeled instances for that relation.

\subsection{Favorable vs.\ Unfavorable
Properties of \cn}

We have investigated several variations of a relation classification model, each variation designed to mitigate some particular feature of \cn~relations. Analysis of these models have shown what impact each has on the model performance, and which 
issues we could address and which we could not. 
One of the issues was the length of the arguments. Using an RNN that can encode such sequences of various lengths did not lead to consistent improvements for relations with long arguments. The classifier still performs best on relations with short arguments. However, we do obtain overall better results with RNN encoding of arguments.

Another issue was the heterogeneity of relations in terms of the semantic or phrasal type of their arguments. The analysis has shown that indeed such relations suffer during classification, but individual tuning of the threshold partly helps.

One of the most striking challenges posed by the \cn~relation inventory remains the observed relation ambiguity. Here, our analysis matches our hypothesis, which was that multi-relation instances are harder to classify than relations for which we rarely find relation instances which co-occur with other relation labels. We further find that individual threshold tuning helps improving classification performance, especially for relations which are harder to classify and are characterized by low F1 scores. These are again exactly the relations which usually show other challenging properties including relation ambiguity, long arguments, and inner-relation diversity regarding concept and phrasal types. 

\subsection{Impact of Missing Edges} 

The ambiguity of \cn~relations combined with the incompleteness of the resource pose challenges for evaluating the performance of a model. A classification decision marked as false positive could in fact be valid. This issue penalizes single-label and multi-label classifiers differently: a single-label classifier is not allowed to predict multiple labels, while a multi-label classifier will learn from potentially false negatives and depending on the distribution of the data could learn to over-predict. To investigate to what degree this issue impacts the results of our model, we manually annotate a small sample of the test data and compare it to the gold standard.

\paragraph{Annotation Experiment.} 
We performed a small
annotation experiment in which we manually control a
subset of 200 instances 
from our test set 
for missing edges. Our sample consists of concept pairs which are related with one of the 14 relations in \textsc{Cn-Omcs-14}, and we want to investigate if another, additional relation holds between the two concepts. We therefore present the concept pair and a randomly sampled relation from our relation set (excluding the gold label) to two annotators without showing the gold label. We ask them if the relation applies or not, and they are also allowed to assign \textit{Not Sure} as a third option. The annotators agreed in 178 of 200 instances (91\%). The annotations are merged by a third expert annotator. In the final gold version 18 (9\%) of the instances are labelled as applicable  (e.g.\ \tr{cook dinner}{HasPrerequisite}{turn on stove}), while 176 (88\%) don't apply according to the annotators (e.g.\ \tr{coffee}{HasSubevent}{popular drink}). 
According to this small annotated subset, we conclude that a lower bound of almost 9\% of the predictions could be penalized due to incompleteness of the \cn~resource or our extracted subset, respectively. Of course this has to be verified in an annotation experiment of a larger scale. 




%% file: tables/relation_set_cn2000.tex
\begin{table}[t]
\centering
\small{
\scalebox{0.8}{
\centering
\begin{tabular}{lccc|lcccc}
\toprule
&multi word& words/term & multi-label &&multi word& words/term & multi-label\\

&terms (\%) & (avg) &  rel.(\%)  &&terms (\%) & (avg) &  rel.(\%)  \\ 
\midrule
\textsc{IsA} & 43.43  & 1.72  &  ~1.70 & \textsc{HasA} & 55.37  & 1.88  &   ~0.46
\\
\textsc{AtLocation} &  36.02  & 1.43  &  ~1.86  &
\textsc{HasProperty} &40.21  & 1.75  &  ~0.62 
 \\
\textsc{UsedFor} & 64.38  & 1.90  &  ~3.25  &
\textsc{CapableOf}& 55.91  & 1.83  &  ~2.86  
 \\
\textsc{ReceivesAction}& 55.91  & 2.21  &   ~0.15  &
\textsc{CausesDesire} & 74.33  & 2.09  &  ~0.23  
\\
\textsc{Desires} & 40.11  & 1.68  &   ~2.01  &
\textsc{MotivatedByGoal} & 78.46  & 2.16  &  ~5.19  
\\
\textsc{HasPrerequisite}& 77.88  & 2.15  &  13.70  &
\textsc{HasFirstSubevent} & 84.78  & 2.30  &  10.84  
\\
\textsc{HasSubevent}& 79.23  & 2.20  &  14.94  &
\textsc{Causes} &72.11  & 2.04  &  ~4.18  
\\
\hline
\textsc{Other} & 53.49  & 1.82  &   ~9.52  &
\textsc{Random} & 55.21  & 1.84  &  ~~ 0 
\\
\bottomrule
\end{tabular}
}
\caption{Relation statistics on \textsc{Cn-Omcs-14}. Results for all relations (Ow-1): 60.01 \% of MW terms, 1.94  (average number of words/term), and 4.77 \%  of relation instances with multiple labels. 
}
\label{tbl:statistics_cn2000}
}
\vspace{-0.1in}
\end{table}

%% file: 7Conclusion.tex
\section{Conclusion}
\label{sec:conclusion}
In this paper we investigated several variations of a multi-label neural relational classifier for \cn~relations. Each variation  was designed to account for specific properties of \cn. An in-depth study revealed specific characteristics that can make \cn~relation classification difficult: 
several distinct relation types may hold for a given concept pair; 
some relations have heterogeneous arguments; and many concepts are expressed through multi-word terms. 
In light of these challenges posed by the specific properties of \cn, we design a multi-label classification model which uses RNNs for representing multi-word arguments and individually tuned thresholds for improving model performance, especially for relations with unfavorable properties such as long arguments, relation ambiguity and inner-relation diversity. Our best performing model achieves F1 scores of 68 in an open world and 71 in a closed world setting. The analysis of the results in different configurations shows that the design decisions driven by multi-word representations and threshold tuning improved the overall classification performance, and that our model is able to tackle 
specific properties of \cn. Yet, some challenges could not be resolved 
and need to be addressed in future work. In particular this concerns relation ambiguity and heterogeneity of relation arguments. 
The observed co-occurences of relations could be deployed for targeting relation ambiguity by building a meta classifier which learns which relations can or cannot occur together.

In future work, we plan to use the multi-label classification system proposed in this paper for enriching \cn~by predicting relations between concepts which are not yet linked in the network. Our investigation can further inform and caution the community on both the usefulness and the flaws of this resource and guide future work on using \cn.\\
